\documentclass[runningheads]{llncs}

\usepackage{eccv}

\usepackage{eccvabbrv}

\usepackage{graphicx}
\usepackage{booktabs}

\usepackage[accsupp]{axessibility}  

\usepackage[pagebackref,breaklinks,colorlinks,citecolor=eccvblue]{hyperref}

\usepackage{orcidlink}
\usepackage{microtype}
\usepackage{subcaption}
\usepackage{amssymb}    
\usepackage[table]{xcolor}     
\usepackage{colortbl}
\usepackage{xcolor} 
\usepackage{tikz}
\usetikzlibrary{tikzmark}
\usepackage{colortbl}   
\usepackage{tikz} 
\usepackage{multirow}
\usetikzlibrary{tikzmark, arrows.meta, calc}
\usepackage{array}            
\newcommand{\cmarkNoCoT}{\cmark^{\text{\kern-2pt\scriptsize$-$}}}

\usetikzlibrary{tikzmark,calc}
\usepackage{amsmath}
\usepackage{mathtools}
\usepackage{tabularx}
\usepackage[most,skins,theorems]{tcolorbox}
\usepackage{tikz}
\usepackage{framed}
\usepackage{pifont}
\usepackage{hyperref}
\usepackage{framed}
\usepackage{ragged2e}

\usepackage[most]{tcolorbox}

\usepackage[table]{xcolor}
\usepackage{mathtools}
\usepackage{tabularx}
\usepackage[most,skins,theorems]{tcolorbox}
\usepackage{xcolor} 
\usepackage{multirow} 
\usepackage[utf8]{inputenc}   
\usepackage[table]{xcolor}
\usepackage{array}
\usepackage{makecell}
\usepackage{pifont}
\usepackage{xcolor} 
\usepackage[table]{xcolor} 
\usepackage{colortbl}
\usepackage{multirow}
\definecolor{myemerald}{rgb}{0.753, 0.898, 0.804}
\definecolor{mylightgreen}{rgb}{0.894, 0.933, 0.745}
\definecolor{myyellow}{rgb}{0.996, 0.972, 0.780}
\definecolor{headergray}{gray}{0.92}
\definecolor{rowblue}{rgb}{0.94, 0.97, 1.0} 
\definecolor{darkred}{rgb}{0.75, 0, 0}

\newcommand{\firstc}{\cellcolor{myemerald!100}}
\newcommand{\secondc}{\cellcolor{mylightgreen!100}} 
\newcommand{\thirdc}{\cellcolor{myyellow!100}}

\newcommand{\cmark}{\textcolor{green!60!black}{\ding{51}}}
\definecolor{backred}{rgb}{1,0.8,0.8}

\begin{document}

\title{Semantic-Geometric Dual Compression: Training-Free Visual Token Reduction for Ultra-High-Resolution  Remote Sensing Understanding} 

\titlerunning{Preprint}

\author{Yueying Li\inst{1}\thanks{Equal contribution} \and
Fengxiang Wang\inst{1}\thanks{Equal contribution} \and
Yan Li\inst{1} \and
Mingshuo Chen\inst{2} \and
Mengying Zhao\inst{1} \and
Long Lan\inst{1}\thanks{Corresponding author}}

\authorrunning{Y. Li and F. Wang et al.}


\institute{College of Computer Science and Technology, National University of Defense Technology, \\ Changsha 410073, China \and
School of Computer Science, Beijing University of Posts and Telecommunications, \\ Beijing 100876, China \\
\email{long.lan@nudt.edu.cn}} 

\maketitle

\begin{abstract}
 Multimodal Large Language Models (MLLMs) have demonstrated immense potential in Earth observation. However, the massive visual tokens generated when processing Ultra-High-Resolution (UHR) imagery introduce prohibitive computational overhead, severely bottlenecking their inference efficiency. Existing visual token compression methods predominantly adopt static and uniform compression strategies, neglecting the inherent "Semantic-Geometric Duality" in remote sensing interpretation tasks. Specifically, object semantic tasks focus on the abstract semantics of objects and benefit from aggressive background pruning, whereas scene geometric tasks critically rely on the integrity of spatial topology. To address this challenge, we propose DualComp, a task-adaptive dual-stream token compression framework. Dynamically guided by a lightweight pre-trained router, DualComp decouples feature processing into two dedicated pathways. In the object semantic stream, the Spatially-Contiguous Semantic Aggregator (SCSA) utilizes size-adaptive clustering to aggregates redundant background while protecting small object. In the scene geometric stream, the Instruction-Guided Structure Recoverer (IGSR) introduces a greedy path-tracing topology completion mechanism to reconstruct spatial skeletons. 
 Experiments on the UHR remote sensing benchmark XLRS-Bench demonstrate that DualComp accomplishes high-fidelity remote sensing interpretation at an exceptionally low computational cost, achieving simultaneous improvements in both efficiency and accuracy. 
  \keywords{UHR Remote Sensing \and Visual Token Compression \and Semantic--Geometric Duality}

\end{abstract}

\section{Introduction}

Recent advances in multimodal large language models (MLLMs)~\cite{gemini,gpt-4,llama,minigpt,qwenvl} have substantially improved visual understanding and reasoning, and have also accelerated progress in Earth science applications involving remote sensing data~\cite{geochat,lhrs,rsgpt}. At the same time, ultra-high-resolution (UHR) remote sensing imagery has greatly expanded the observable detail available for Earth observation: it captures not only large geographic regions, but also fine-grained cues about human activities and natural environments~\cite{dota,semi}. However, UHR imagery also produces extremely long visual token sequences in MLLMs, which significantly increase inference cost, memory usage, and latency~\cite{llavauhd3}. As a result, efficient processing of UHR remote sensing imagery has become a major bottleneck for practical remote sensing MLLMs.

To mitigate the burden of long visual sequences in MLLMs~\cite{llavanext,fargpt-4v}, existing MLLM approaches mainly follow three directions. In the general domain, recent methods typically reduce visual tokens through token merging on the vision side~\cite{prumerge,visionzip,vscan}, token pruning during LLM decoding~\cite{fastv,pdrop,FitPrune,atp-llava}, or instruction-aware compression guided by cross-modal relevance~\cite{instructblip,mplug,flamingo,glimpse}. 
These strategies have also inspired early remote sensing adaptations~\cite{geollava8k,lrsvqa}, where representative systems such as GeoLLaVA-8K~\cite{geollava8k} introduce visual token compression, ZoomEarth~\cite{zoomearth} tackles UHR remote sensing through active perception. 
Despite their differences, these approaches largely treat efficiency as a problem of reducing or reallocating visual inputs, yet still lack an explicit task-adaptive principle for deciding what should be compressed and what should be preserved in remote sensing interpretation. This limitation becomes particularly critical in UHR remote sensing, where the value of visual evidence is highly task-dependent.

\begin{figure}[htbp]
\centering
\vspace{-2mm}
\includegraphics[width=0.99\columnwidth]{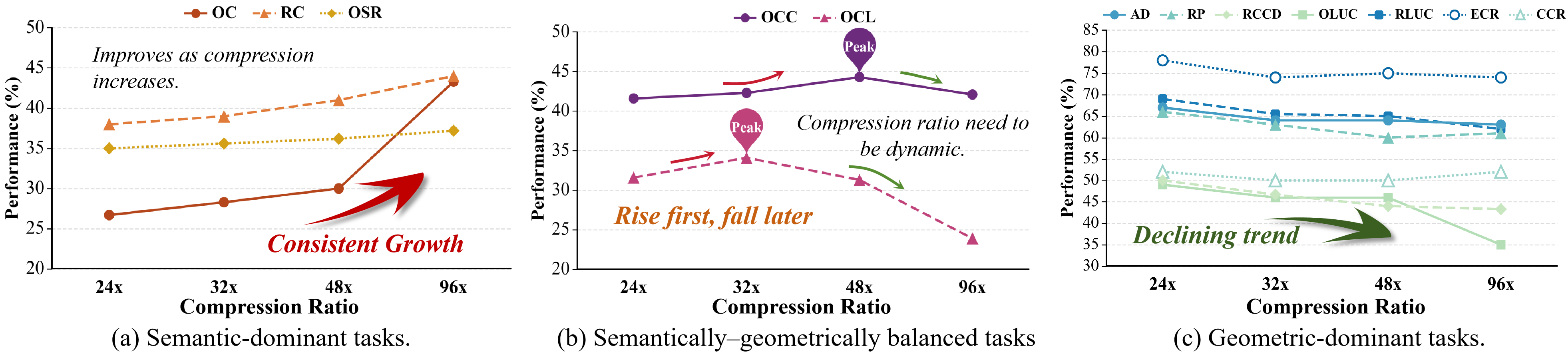}
\caption{Pilot study reveals semantic–geometric duality under token compression. We evaluate a UHR remote-sensing MLLM across multiple compression ratios and report performance changes for three representative groups of subtasks: (a) semantic-dominant improves as compression increases; (b) balanced peaks at a moderate ratio; (c) geometric-dominant drops under strong compression. This highlights that background tokens can be either redundant noise or indispensable geometric evidence, depending on task intent.
}
\vspace{-3mm}
\label{fig:pilot}
\end{figure}

\begin{figure*}[t]

\centering
\includegraphics[width=0.99\textwidth]{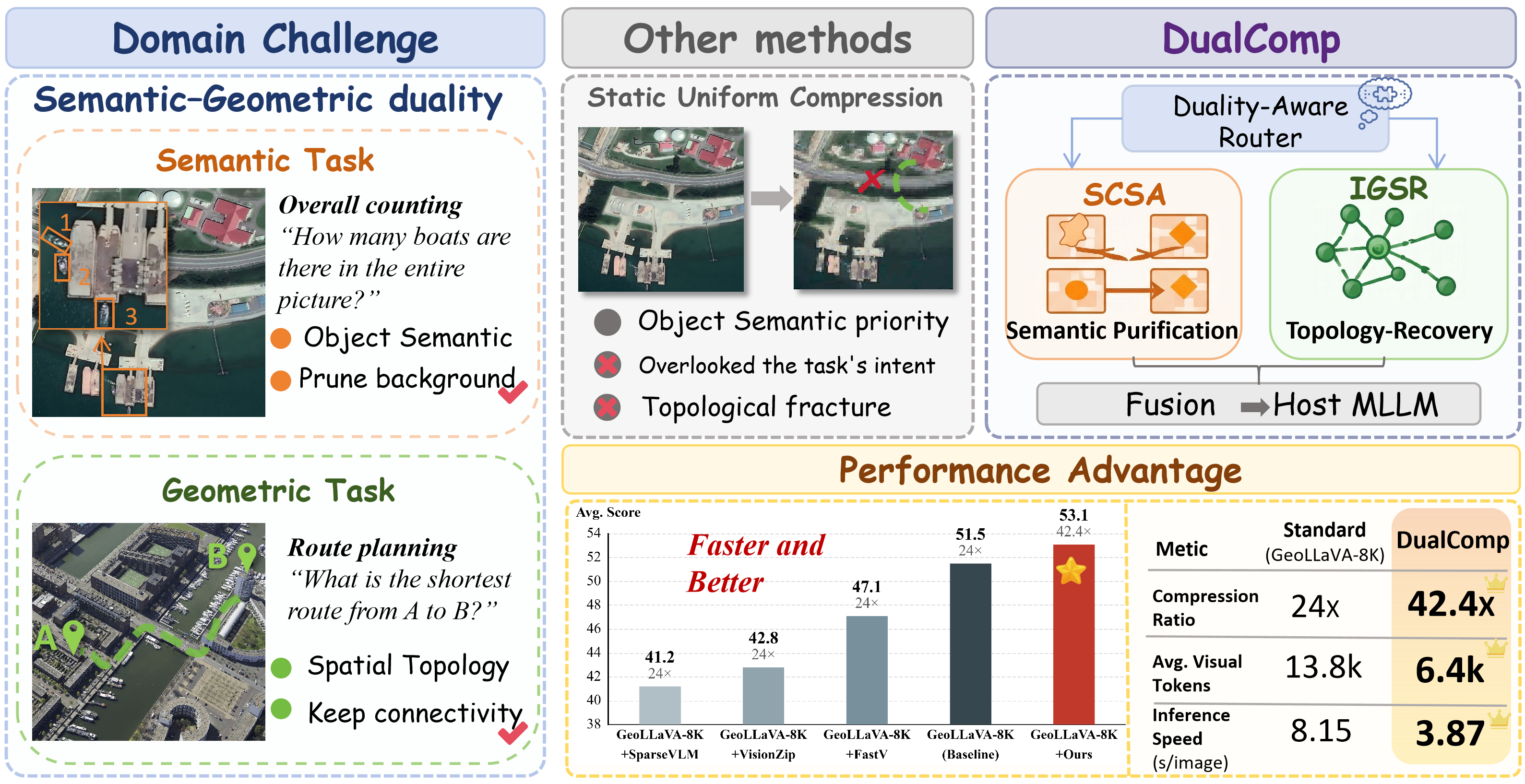}
\caption{DualComp achieves performance improvements under high compression ratios through semantic–geometric dual-stream adaptive compression. We reveal a pervasive semantic–geometric duality in UHR remote sensing MLLMs.
}
\label{fig:intro}
\vspace{-3mm}
\end{figure*}

A key observation of this work is that UHR remote-sensing interpretation exhibits a pronounced \emph{semantic--geometric duality}. We validate this via a pilot study on a representative UHR remote-sensing MLLM by sweeping compression strengths and tracking subtask performance as background-redundant tokens are progressively removed. As shown in \cref{fig:pilot}, the curves consistently fall into three regimes: \textbf{(i)} semantic-dominant tasks improve with stronger compression (\emph{semantic purification}); \textbf{(ii)} balanced tasks peak at moderate compression; and \textbf{(iii)} geometric-dominant tasks degrade under strong compression, indicating the necessity of preserving context and topology. Overall, remote-sensing interpretation exhibits a pronounced \emph{semantic--geometric duality}: semantic understanding relies more on object-related foreground evidence, whereas geometric reasoning requires preserving sufficient background context and topology---therefore, the utility of the same visual token can dynamically switch between the two with task intent, making token importance inherently task-dependent.

Motivated by this observation, we argue that remote sensing MLLMs should move from passive information reduction to \emph{task-adaptive information scheduling}. To this end, we propose \textbf{DualComp} (Duality-Aware Token Compression), a plug-and-play dynamic visual token compression framework designed for UHR remote sensing. DualComp decomposes compression into two complementary pathways: a \emph{semantic stream} that preserves object-level semantic fidelity, and a \emph{geometric stream} that preserves scene-level structural fidelity. Specifically, we design a lightweight parasitic router that infers task intent from the user instruction and dynamically allocates token budgets between semantic and geometric evidence. The semantic stream is implemented by a Spatially-Contiguous Semantic Aggregator (SCSA), which compresses redundant background while preserving critical object-level information, and the geometric stream is implemented by an Instruction-Guided Structure Recoverer (IGSR), which preserves and reconstructs task-relevant geometric structures. The compression modules are training-free, while the lightweight router is pretrained offline and frozen at inference, enabling plug-and-play token reduction with no updates to the host MLLM weights and improved efficiency and performance under high compression.

In summary, our main contributions are as follows:
\begin{enumerate}
    \item We identify and quantitatively validate a pervasive \emph{semantic--geometric duality} in UHR remote sensing MLLMs, and show that existing unified and static visual token compression strategies are fundamentally mismatched to the heterogeneous demands of remote sensing tasks.
    
    \item We propose \textbf{DualComp}, a task-intent-aware dynamic visual token compression framework that explicitly models task demands through a lightweight router and adaptively preserves semantic and geometric evidence via the dual-stream design of SCSA and IGSR.
    
    \item We develop a plug-and-play compression pipeline that significantly reduces the inference cost of UHR remote sensing imagery while consistently improving overall task performance, demonstrating the effectiveness of task-intent-aware compression for remote sensing MLLMs.
\end{enumerate}

\section{Related Work}

\subsection{Visual Token Compression for MLLMs}
As MLLMs become increasingly capable of processing high-resolution images, the visual token sequence grows rapidly, while the quadratic complexity of Transformer self-attention~\cite{attention} makes inference cost a critical bottleneck. To address this issue, one line of work converts high-resolution patch grids into more compact representations through explicit downsampling~\cite{llava-onevision}, spatial transformations~\cite{internvl2,internvl3,qwen25vl,qwen3vl}, or lightweight projections~\cite{flamingo,minigpt,instructblip,mplug} including LLaVA-OneVision~\cite{llava-onevision}, Qwen2.5-VL~\cite{qwen25vl}, InternVL2~\cite{internvl2}, Blip-2~\cite{blip2}, and Honeybee~\cite{honeybee}, but these approaches usually require architectural modifications and additional training overhead. Another line of research explores training-free token reduction~\cite{fastv, cls,sparsevlm,visionzip,vscan}, primarily on the visual encoder side and the LLM decoding side, such as ToMe~\cite{tome} and  VisionZip~\cite{visionzip}. During decoding, methods such as FastV~\cite{fastv}, SparseVLM~\cite{sparsevlm}, PyramidDrop~\cite{pdrop}, and ATP-LLaVA~\cite{atp} accelerate inference by pruning tokens at specific layers or in staged manners based on attention signals. In addition, approaches such as SparseVLM~\cite{sparsevlm} and AdaFV~\cite{adafv} explicitly leverage text queries for cross-modal matching and selection, while VisionTrim~\cite{visiontrim} also uses text guidance to perform context-aware token merging within a more complete MLLM pipeline. 
Overall, most of these studies treat token compression as a single-dimensional problem of importance estimation, yet they often overlook the semantic–geometric duality inherent in ultra-high-resolution remote sensing scenes.

\subsection{MLLMs in UHR Remote Sensing Understanding}
General-purpose multimodal large language models (MLLMs), such as LLaVA~\cite{llava} and Intern-S1~\cite{interns1}, have demonstrated strong visual understanding and instruction-following capabilities, driving the rapid development of remote sensing MLLMs. Early efforts mainly focused on aligning and adapting remote sensing visual encoders to general-purpose LLMs, such as RSGPT~\cite{rsgpt}, SkyEyeGPT~\cite{skyeyegpt}, GeoChat~\cite{geochat}, and EarthGPT~\cite{earthgpt}. Subsequent studies further improved data construction, alignment strategies, and instruction-following ability, including VHM~\cite{VHM}, RS-CapRet~\cite{RScapret}, EarthMarker~\cite{Earthmarker}, LHRS-Bot-Nova~\cite{LHRSBotNova}, RSUniVLM~\cite{RSUniVLM}, EarthMind~\cite{EarthMind}, EarthDial~\cite{Earthdial}, RingMoGPT~\cite{Ringmogpt}, and EarthVL~\cite{EarthVL}.However, in ultra-high-resolution (UHR) remote sensing scenarios, these models often struggle to accurately localize task-relevant fine-grained regions within vast pixel spaces. To address this challenge, some studies have introduced visual token compression and selection strategies within MLLMs, such as GeoLLaVA-8K~\cite{geollava8k}, ImageRAG~\cite{imagerag}, and RFM~\cite{lrsvqa}. 
Another line of work shifts toward a tool-use paradigm, as exemplified by ZoomEarth~\cite{zoomearth} and VICoT-Agent~\cite{vicot}, which acquire local details through chain-of-thought-driven multi-step zooming or retrieval. Yet, these approaches often require additional interaction rounds and incur substantial token overhead, making them still constrained by the trade-off between efficiency and scalability in UHR settings.


\section{Method}
\subsection{Preliminary Analysis}

\noindent\textbf{Visual Token Compression in UHR MLLMs.} When MLLMs process UHR remote sensing images, AnyRes-style multi-cropping~\cite{llavanext} and dense patch-based representations often produce a massive number of visual tokens, leading to substantial memory and latency overhead.
Existing static compression baselines typically shorten the visual sequence by uniformly clustering and merging local visual features without modifying the backbone model~\cite{visionzip,sparsevlm,fastv,pdrop,geollava8k}. However, as illustrated in \cref{fig:pilot}, our empirical analysis under extreme compression ratios ($24\times$--$96\times$) reveals a systematic limitation of this \emph{static and uniform} paradigm in UHR remote sensing scenarios: it fails to account for the intrinsic semantic--geometric duality of remote sensing tasks, resulting in clear mismatches across different task intents.

More specifically, UHR remote sensing tasks lie on a semantic--geometric demands. For clarity, we describe three representative regimes:
\begin{itemize}
    \item \textbf{Semantic-dominant tasks.} These tasks rely more heavily on object attributes and instance-level statistics, such as counting, presence detection, and relative relationships between targets. In essence, they focus more on \emph{what is present}. In such cases, large homogeneous background regions often function mainly as noise, and the model relies more on recognizing discrete objects than on preserving precise topology or long-range connectivity.

    \item \textbf{Semantically--geometrically balanced tasks.} These tasks require both object-level semantic cues and fine-grained local structure. Typical examples include object classification and object color recognition, where moderate background compression can reduce irrelevant noise, but excessive compression may also remove discriminative details or context needed for reliable prediction. As a result, these tasks typically favor a balanced preservation of semantic and geometric evidence rather than an extreme compression policy.

    \item \textbf{Geometric-dominant tasks.} These tasks rely more heavily on spatial organization and structural reasoning, including path planning, land-use or functional zone classification, and boundary understanding. In essence, they focus more on \emph{where objects are} and \emph{how spatial structures are organized}. Their success depends heavily on spatial integrity, continuous paths, precise boundaries, and contextual relations across regions. Static compression guided mainly by semantic saliency may therefore discard or over-merge tokens carrying critical structural evidence, causing severe performance degradation.
\end{itemize}
These observations indicate that, in UHR remote sensing interpretation, the set of high-value tokens changes dynamically with task intent, and a single compression ratio is insufficient for all tasks. 

\subsection{Our Approach}

\noindent To address the compression mismatch caused by the semantic--geometric duality of ultra-high-resolution (UHR) remote sensing tasks, we propose DualComp, a task-adaptive dual-stream visual token compression framework. DualComp formulates visual token compression as an instruction-conditioned adaptive allocation problem and implements it through a dynamic \emph{Perception--Decision--Execution} pipeline.

\begin{figure*}[t]
\centering
\includegraphics[width=0.99\textwidth]{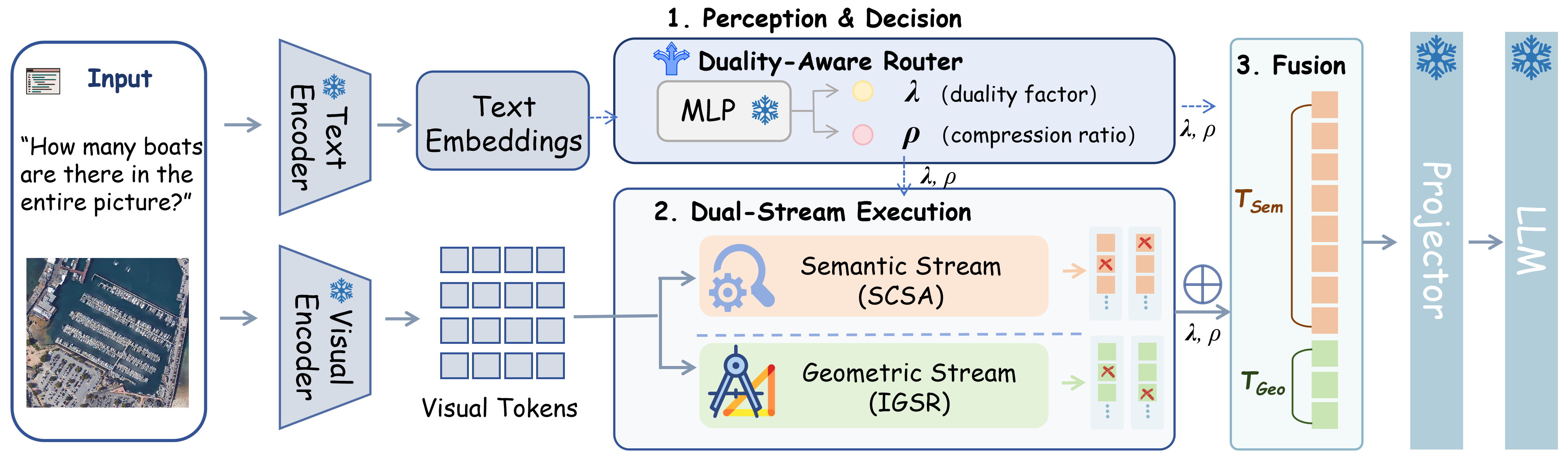}
\caption{\textbf{Overview of DualComp.} Text-conditioned routing outputs $(\lambda,\rho)$, steering SCSA (semantic) and IGSR (geometric) token reduction and fusion.}
\label{fig:overview}
\vspace{-2mm}
\end{figure*}

\noindent As illustrated in \cref{fig:overview}, the overall workflow of DualComp consists of two stages:

\begin{enumerate}
    \item \textbf{Perception and Decision: Duality-Aware Router.}  
    A lightweight \emph{duality-aware router} parses the textual instruction and predicts a task-specific compression policy, including the overall retention strength and the relative preference between semantic and geometric evidence. In this way, the router determines both how aggressively the visual tokens should be compressed and how the retained budget should be scheduled across the two streams.

    \item \textbf{Dual-Stream Execution and Fusion: SCSA and IGSR.}  
    In the execution stage, DualComp decomposes visual token compression into two complementary streams. The semantic stream uses the Spatially-Contiguous Semantic Aggregator (SCSA) to compress redundant background regions while preserving object-level semantics. The geometric stream uses the Instruction-Guided Structure Recoverer (IGSR) to preserve connectivity, boundaries, and other geometry-critical structures. The two streams are then fused by simple feature-level operations and directly fed into the MLLM without additional learnable projection layers.
\end{enumerate}

\subsubsection{Duality-Aware Router.}

The Router serves as the control center of DualComp. Given a textual instruction, it predicts two task-specific control variables: a duality factor $\lambda \in [0,1]$ and an overall compression ratio $\rho \in [\rho_{\min}, 1]$. Here, a smaller $\lambda$ indicates a stronger semantic preference, while a larger $\lambda$ indicates a stronger geometric preference, and $\rho$ controls the overall retention strength. Given the initial upper bound of visual tokens $N_{\max}$, the framework determines the total retention budget as $n_{\text{keep}} = N_{\max}\rho$, and then allocates it to the semantic and geometric streams as $n_{\text{sem}} = n_{\text{keep}}(1-\lambda)$ and $n_{\text{geo}} = n_{\text{keep}}\lambda$.

To minimize overhead, we adopt a parasitic design that attaches the Router directly to the text embedding output of the host MLLM. Text features are compressed into a compact instruction representation and fed into a shared multilayer perceptron (MLP) with two independent Sigmoid heads, which predict $\lambda$ and $\rho$, respectively. The Router contains only about $1$M parameters, remains frozen during inference, and preserves the plug-and-play nature of DualComp.

Since task duality preference lacks manually annotated labels, we construct an offline label generation pipeline with a dual-perspective annotation scheme for Router training. Specifically, we obtain $\lambda_{\text{llm}}$ through Likert-scale probing with the host LLM and derive $\lambda_{\text{rule}}$ from expert-designed linguistic rules. The final supervision signal is defined as $\lambda_{\text{gt}} = \alpha \cdot \lambda_{\text{llm}} + (1-\alpha) \cdot \lambda_{\text{rule}}$.

\subsubsection{Spatially-Contiguous Semantic Aggregator (SCSA)}

\begin{figure*}[t]
\centering
\includegraphics[width=0.99\textwidth]{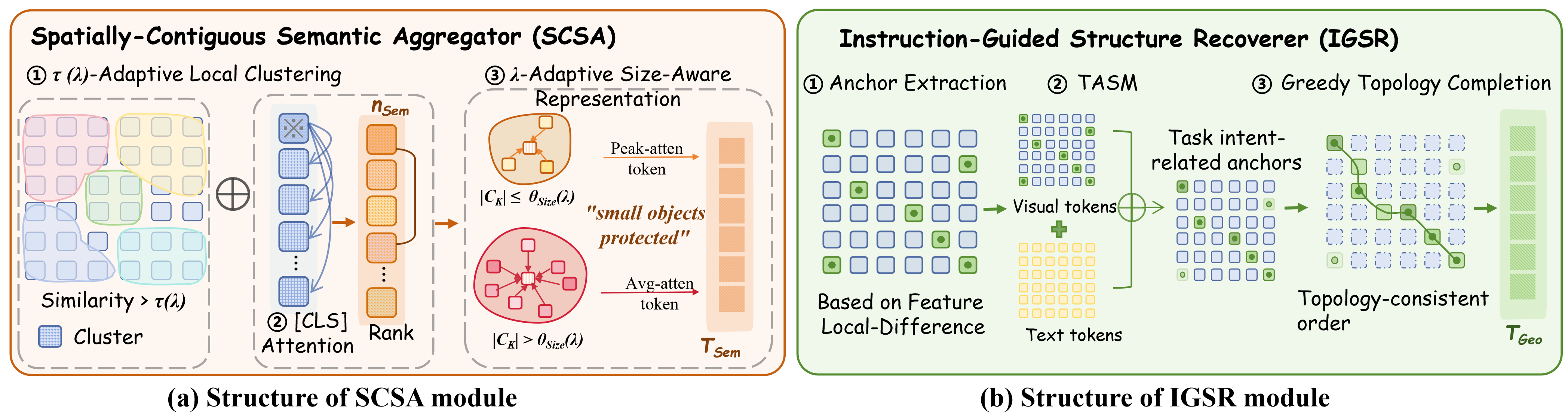}
\caption{\textbf{Two compression modules in DualComp.} (a) SCSA aggregates redundant background while preserving semantic cues. (b) IGSR recovers topological cues via instruction-guided completion.}

\label{fig:detail}
\vspace{-2mm}
\end{figure*}

To reduce background redundancy when the task emphasizes object semantics, we introduce SCSA as the semantic stream of DualComp. Given the Router-allocated budget $n_{\text{sem}}$, SCSA performs training-free compression by aggregating homogeneous background regions while preserving instance-level information for small objects. As illustrated in \cref{fig:detail}(a), it consists of three stages.
\paragraph{\textbf{1. $\tau(\lambda)$-Adaptive Local Clustering.}}
SCSA first groups locally similar tokens into spatially contiguous clusters to compress redundant background regions. We compute cosine similarity using frozen CLIP~\cite{clip} visual features and introduce a dynamic threshold $\tau(\lambda)$ controlled by the duality factor $\lambda$. For a token $i$, merging is allowed only when the maximum similarity within its local neighborhood exceeds the threshold:
\begin{equation}
    \mathrm{parent}(i) =
    \begin{cases}
    \arg\max_{j \in \mathcal{N}(i),\, j<i} \cos(f_i,f_j), & \text{if } \max \cos > \tau(\lambda), \\
    i, & \text{otherwise}.
    \end{cases}
    \label{eq:parent_def}
\end{equation}
Here, $\tau(\lambda)$ is a monotonically increasing mapping, encouraging more aggressive aggregation when $\lambda$ is small and preserving finer local granularity as $\lambda$ increases. Detailed clustering implementation is provided in the appendix.

\paragraph{\textbf{2. CLS Attention-Based Cluster Scoring.}}
Given the resulting clusters $\{c_1, c_2, \dots, c_K\}$, SCSA selects the top $n_{\text{sem}}$ clusters according to semantic importance. We use the \texttt{[CLS]}-to-patch attention in a pretrained ViT as a measure of global semantic relevance~\cite{cls}:
\begin{equation}
    a_{[\mathrm{CLS}]} = \mathrm{Softmax}\left(\frac{q_{[\mathrm{CLS}]}K_v^T}{\sqrt{d}}\right)
    \label{eq:attn_weights}
\end{equation}
Cluster importance is then computed by cumulative attention:
\begin{equation}
    \mathrm{Importance}(c_k) = \sum_{i \in c_k} a_{[\mathrm{CLS}],i}
    \label{eq:importance}
\end{equation}
This yields a compact set of semantically important clusters for subsequent representation.

\paragraph{\textbf{3. $\lambda$-Adaptive Size-Aware Representation.}}
We further apply a size-aware representation strategy controlled by a dynamic threshold $\theta_{\text{size}}(\lambda)$. For small clusters ($|c_k| \leq \theta_{\text{size}}$), we retain the original token with the highest \texttt{[CLS]} attention. For large clusters ($|c_k| > \theta_{\text{size}}$), we use an attention-weighted average as a summary token. Together, these steps enable SCSA to balance background compression and small-object preservation under different task intents, under the unified control of the Router's $\lambda$.

\subsubsection{Instruction-Guided Structure Recoverer (IGSR)}

To preserve structural evidence for scene geometric tasks, DualComp introduces IGSR to utilize the Router-allocated budget $n_{\text{geo}}$. IGSR performs topological reconstruction on the compressed feature grid to recover connectivity, boundaries, and other geometry-critical structures. As illustrated in \cref{fig:detail}(b), it is a parameter-free module consisting of three steps.

\paragraph{\textbf{1. Feature Local-Difference Anchor Extraction.}}
IGSR first estimates geometric saliency from local feature variation:
\begin{equation}
    S_{\text{geo}}(i) = \left\|F(i) - \mathrm{AvgPool}_{3\times 3}(F)(i)\right\|_2^2
    \label{eq:s_geo}
\end{equation}
To ensure spatial coverage, IGSR retains the highest-scoring token in each subregion as a structural anchor. Detailed anchor selection is provided in the appendix.

\paragraph{\textbf{2. Text-Aware Structural Modulation.}}
To make reconstruction instruction-aware, IGSR further introduces a text-aware structural modulation signal (TASM). Specifically, we compute a text relevance score $S_{\text{text}}$ using CLIP~\cite{clip} text--vision similarity and use it to modulate the structural cost field:
\begin{equation}
    S_{\text{struct}}(i) = \tilde{S}_{\text{geo}}(i) \cdot \left(1 + \beta \cdot \tilde{S}_{\text{text}}(i)\right)
    \label{eq:s_struct}
\end{equation}
where $\tilde{S}_{\text{geo}}$ and $\tilde{S}_{\text{text}}$ are normalized scores. This biases reconstruction toward instruction-relevant structures, while naturally reducing to purely geometric reconstruction when the text signal is weak.

\paragraph{\textbf{3. Parallel Greedy Topology Completion.}}
Given structural anchor pairs $(A_i, A_{i+1})$, IGSR recovers their connecting path on the discrete feature grid through a fully parallel \emph{Greedy Path Tracing} strategy. At each step, the next node is selected from the n-neighborhood $\mathcal{N}_n(v_t)$ as the candidate with the highest structural cost, under a Chebyshev-distance-decreasing constraint:
\begin{equation}
    v_{t+1} = \arg\max_{u \in \mathcal{N}_n(v_t)} S_{\text{struct}}(u), \quad  \text{s.t. } D_{\text{Chebyshev}}(u, A_{i+1}) < D_{\text{Chebyshev}}(v_t, A_{i+1})
    \label{eq:path_update}
\end{equation}
Implemented with parallel tensor operations, this step restores structural connectivity with minimal additional overhead.

Together, these steps restore critical structural evidence, allowing IGSR to provide the geometric stream of DualComp with efficient structural fidelity.

\subsubsection{Dual-Stream Fusion and Sequence Unrolling}

After SCSA produces the condensed semantic features $T_{\text{sem}}$ and IGSR reconstructs the geometrically continuous features $T_{\text{geo}}$, DualComp fuses the two streams and feeds them into the host MLLM. To preserve plug-and-play compatibility, no additional projection or normalization layers are introduced. Instead, we adopt a strategy based on $\lambda$-guided fusion and topological sequence unrolling.

\paragraph{\textbf{1. $\lambda$-Encoded Dual-Stream Fusion.}}

We perform fusion through $\lambda$-based weighted concatenation, $T_{\text{fused}}=\mathrm{Concat}\big((1-\lambda)\,T_{\text{sem}},\,\lambda\,T_{\text{geo}}\big)$, which injects task intent at both the token-allocation and feature-magnitude levels: $(1-\lambda):\lambda$ controls the semantic--geometric token ratio, while the weights softly suppress the non-dominant stream under strong task preference.

\paragraph{\textbf{2. Topological Sequence Unrolling.}}
Under extreme compression, token dropping and clustering may disrupt the original 2D spatial structure. DualComp alleviates this issue through \emph{topological sequence unrolling}, where IGSR outputs a 1D geometric token sequence $T_{\text{geo}}$ ordered by spatial connectivity. When fed into the host LLM, its native relative positional encoding (e.g., RoPE) can capture local continuity along this sequence. Without modifying positional encoding parameters, the host LLM can therefore leverage its autoregressive context modeling to reason over underlying 2D connectivity.

\section{Experiments}
\subsection{Experimental Setup}
\subsubsection{Implementation Details and Benchmark.}
We deploy DualComp on a UHR remote-sensing-specific MLLM and further validate its generality on Qwen2.5-VL~\cite{qwen25vl}. For fair comparison, both experimental tracks follow the default settings of their respective original papers under the same evaluation protocol. 
We adopt XLRS-Bench~\cite{xlrsbench}, currently the largest and highest-resolution multimodal benchmark for remote sensing, as our primary evaluation platform. It contains ultra-high-resolution remote sensing images and covers 13 fine-grained VQA subtasks, spanning the full spectrum from object-semantic-oriented tasks (e.g., counting and object classification) to scene-geometric-oriented tasks (e.g., route planning and anomaly detection). 

\subsection{Main Results}

\begin{table*}[t!]
\footnotesize
\centering
\caption{\textbf{Experimental results on the perception and reasoning dimensions on XLRS-Bench.} 
The sub-tasks include \textbf{Perception}: Overall/Regional Counting (OC/RC), Overall/Regional Land Use Classification (OLUC/RLUC), Object Classification (OCC), Object Color (OCL), Object Motion State (OMS), and Object Spatial Relationship (OSR); and \textbf{Reasoning}: Anomaly Detection and Interpretation (AD), Environmental Condition Reasoning (ECR), Route Planning (RP), Regional Counting with Change Detection (RCCD), and Counting with Complex Reasoning (CCR). 
`Avg.' represents the macro average accuracy across sub-tasks. 
Rankings are calculated column-wise: \colorbox{myemerald}{green} (1st, bold), \colorbox{mylightgreen}{light green} (2nd), and \colorbox{myyellow}{yellow} (3rd). Tie scores are treated identically.}
\label{tab:main-result}
\resizebox{\textwidth}{!}{%
\begin{tabular}{l|c|cccccccc|ccccc|c}
\toprule
\rowcolor[gray]{.9} \textbf{Method} & \textbf{Compression} & \multicolumn{8}{c|}{\textbf{Perception}} & \multicolumn{5}{c|}{\textbf{Reasoning}} & \textbf{Avg.} \\
\midrule
\rowcolor[gray]{.9} \textbf{Sub-tasks (L-3 Capability)} & & \textbf{OC} & \textbf{RC} & \textbf{OSR} & \textbf{OLUC} & \textbf{RLUC} & \textbf{OCC} & \textbf{OCL} & \textbf{OMS} & \textbf{AD} & \textbf{ECR} & \textbf{RP} & \textbf{RCCD} & \textbf{CCR} & \\
\midrule
\multicolumn{16}{l}{\textit{\textcolor{gray}{Closed-source MLLMs}}} \\
Claude 3.7 Sonnet~\cite{claude} & - & 27.6 & 22.7 & 27.6 & 17.4 & 68.4 & 30.5 & 29.9 & 63.6 & 64.8 & 78.4 & 34.5 & 27.8 & 32.6 & 40.5 \\
Gemini 2.5 Pro~\cite{gemini25} & - & - & - & - & - & - & - & - & - & - & - & - & - & - & 45.2 \\
GPT-5.2~\cite{gpt52} & - & 30.0 & 37.0 & 34.0 & 17.0 & 70.5 & 43.0 & \thirdc 41.4 & \firstc \textbf{68.3} & \secondc 74.0 & 76.0 & 52.0 & 36.7 & 38.0 & 47.5 \\
\midrule
\multicolumn{16}{l}{\textit{\textcolor{gray}{Open-source MLLMs}}} \\
LLaVA-Next~\cite{llavanext} & - & 26.7 & 40.0 & 30.0 & 5.0 & 67.0 & 28.8 & 32.8 & 66.7 & 69.0 & 78.0 & 27.0 & 35.0 & 36.0 & 41.7 \\
LLaVA-Next+RFM+DIP~\cite{lrsvqa} & $4\times$ & \thirdc 36.7 & 41.0 & 25.6 & 2.0 & 54.5 & 33.9 & 34.0 & 53.3 & 70.0 & 76.0 & 24.0 & \firstc \textbf{53.3} & 44.0 & 42.2 \\
InternVL3-8B~\cite{internvl3} & - & \firstc \textbf{40.0} & 39.0 & 25.2 & 10.0 & 71.5 & \thirdc 44.5 & 30.8 & \thirdc 65.0 & \firstc \textbf{77.0} & \secondc 82.0 & 36.0 & 21.7 & \thirdc 50.0 & 45.6 \\
Qwen2-VL-7B~\cite{qwen2vl} & - & 26.7 & 40.0 & 31.8 & 11.0 & 73.0 & 35.9 & 34.6 & 61.7 & 70.0 & \thirdc 81.0 & 35.0 & 46.7 & 48.0 & 45.8 \\
InternVL2.5-8B~\cite{internvl2} & - & \secondc 38.3 & 37.0 & 21.6 & 10.0 & \thirdc 77.0 & 33.4 & 35.5 & \thirdc 65.0 & \thirdc 73.0 & \firstc \textbf{83.0} & 34.0 & \secondc 50.0 & 43.0 & 46.2 \\
Qwen2.5-VL-7B~\cite{qwen25vl} & - & 33.3 & 40.0 & 36.2 & 31.0 & \thirdc 77.0 & 40.6 & 40.5 & \secondc 66.7 & 68.0 & 72.0 & 27.0 & 38.3 & 45.0 & 47.4 \\
Qwen3-VL-8B~\cite{qwen3vl} & - & 21.7 & \firstc \textbf{50.0} & 30.4 & 26.0 & \firstc \textbf{81.5} & \firstc \textbf{46.6} & \firstc \textbf{43.1} & \secondc 66.7 & \secondc 74.0 & 79.0 & 37.0 & 43.3 & \secondc 51.0 & 50.0 \\
Qwen2.5-VL-72B~\cite{qwen25vl} & - & 33.3 & \secondc 47.0 & 34.0 & 39.0 & \secondc 80.0 & \secondc 45.3 & \secondc 42.1 & \thirdc 65.0 & 71.0 & 74.0 & 37.0 & 43.3 & 42.0 & 50.2 \\
Intern-S1-mini~\cite{interns1} & - & - & - & - & - & - & - & - & - & - & - & - & - & - & \secondc 51.6 \\
\midrule
\multicolumn{16}{l}{\textit{\textcolor{gray}{Remote Sensing MLLMs}}} \\
GeoChat~\cite{geochat} & - & 16.7 & 29.0 & 24.2 & 2.0 & 23.0 & 21.1 & 16.8 & 35.0 & 33.0 & 43.0 & 10.0 & - & 21.0 & 22.9 \\
ZoomEarth~\cite{zoomearth} & - & - & - & - & - & - & - & - & - & - & - & - & - & - & 40.2 \\
GeoLLaVA-8K~\cite{geollava8k} & - & 26.7 & 38.0 & 35.0 & \secondc 49.0 & 69.0 & 41.6 & 31.6 & \thirdc 65.0 & 67.0 & 78.0 & \secondc 66.0 & \secondc 50.0 & \firstc \textbf{52.0} & \thirdc 51.5 \\
GeoLLaVA-8K+VisionZip~\cite{visionzip} & $24\times$ & 23.3 & 39.0 & \secondc 38.6 & 37.0 & 49.0 & 44.1 & 30.0 & \thirdc 65.0 & 36.0 & 38.0 & 62.0 & 46.7 & 47.0 & 42.8 \\
GeoLLaVA-8K+FastV~\cite{fastv} & $24\times$ & 23.3 & 41.0 & 37.4 & \thirdc 46.0 & 50.5 & 43.8 & 31.6 & \thirdc 65.0 & 53.0 & 60.0 & \thirdc 65.0 & 46.7 & 49.0 & 47.1 \\
GeoLLaVA-8K+SparseVLM~\cite{sparsevlm} & $24\times$ & 21.7 & 39.0 & \firstc \textbf{38.8} & 38.0 & 32.5 & 33.8 & 26.5 & \thirdc 65.0 & 43.0 & 43.0 & 62.0 & 45.0 & 47.0 & 41.2 \\
\midrule
\rowcolor{rowblue}
\textbf{GeoLLaVA-8K+Ours} & \normalsize  \textcolor{darkred}{$\mathbf{42.4\times}$} & 26.7 & \thirdc 45.0 & 37.4 & \firstc \textbf{53.0} & 69.5 & 43.6 & 34.0 & \thirdc 65.0 & 69.0 & 79.0 & \firstc \textbf{72.0} & 48.3 & 49.0 & \firstc \textbf{53.1} \\
\bottomrule
\end{tabular}%
}
\vspace{-5mm}
\end{table*}

\subsubsection{Overall Performance Comparison.}
\cref{tab:main-result} shows that generic token reduction methods (VisionZip, SparseVLM, and FastV) consistently underperform on UHR remote sensing tasks, indicating that task-agnostic importance heuristics often discard remote-sensing-specific evidence. In contrast, DualComp achieves the best overall accuracy ($53.10\%$), outperforming both generic baselines and the RS-tailored static compression baseline GeoLLaVA-8K ($51.50\%$, +1.6 points). This suggests that task-intent-aware scheduling is more effective than fixed compression policies for balancing semantic and geometric evidence.

The largest gains appear on geometry-sensitive tasks, where performance depends on connectivity and structural integrity. With the geometric stream IGSR, DualComp improves \emph{Route Planning} from $66\%$ to $72\%$ and \emph{Overall Land Use Classification} from $49\%$ to $53\%$, with similar gains on other geometry-heavy reasoning subtasks. This confirms that preserving topology-critical evidence is essential under aggressive compression.

DualComp also remains competitive on semantic-dominant tasks. The semantic stream SCSA improves \emph{Regional Counting} ($38\% \rightarrow 45\%$) and \emph{Object Color} ($31.6\% \rightarrow 34.0\%$), while maintaining comparable performance on most object-centric subtasks. The OMS accuracy remains virtually unchanged across methods, which we attribute to a limitation of the current evaluation set; the same invariance is observed on Qwen-based backbones. A few hybrid tasks still show limited gains or slight drops, suggesting that tasks jointly requiring fine-grained instance cues and broader context remain more sensitive to budget partitioning. Nevertheless, DualComp achieves the best overall performance across the semantic--geometric remote sensing tasks.

\begin{table}[b!]
\centering
\caption{\textbf{Comparison of computational efficiency with different token reduction methods based on GeoLLaVA-8K.}}
\label{tab:efficiency-comparison}
\resizebox{\textwidth}{!}{%
\begin{tabular}{l|c|c|c|c|c}
\toprule
\rowcolor[gray]{.85}
\textbf{Metrics} & \textbf{GeoLLaVA-8K} & \textbf{w/ VisionZip} & \textbf{w/ FastV} & \textbf{w/ SparseVLMs} & \textbf{w/ Ours} \\
\midrule
\rowcolor[gray]{.95}
\textbf{Compression Ratio} & $24\times$ & $24\times$ & $24\times$ & $24\times$ & $\mathbf{42.4\times}$ \\
\textbf{Tokens per Grid} & 24 & 24 & 24 & 24 & \textbf{14.2} \\
\rowcolor[gray]{.95}
\textbf{Avg. Visual Token Volume} & 13.8k & 13.8k & 13.8k & 13.8k & \textbf{6.4k} \\
\textbf{TFLOPs of LLM} & 198.1 & 198.7 & 198.1 & 186.1 & \textbf{99.8} \\
\midrule
\rowcolor[gray]{.95}
\textbf{Inference Speed (s/image)} & 8.15 & 8.41 & 19.56 & 7.84 & \textbf{3.87} \\
\quad\emph{Visual Encoding + Compression} & 4.28 & 5.08 & 8.59 & 4.72 & \textbf{1.52} \\
\rowcolor[gray]{.95}
\quad\emph{LLM Generation} & 3.87 & 3.33 & 10.97 & 3.13 & \textbf{2.25} \\
\midrule
\textbf{Avg. Score} & 51.5\% & 42.75\% & 47.10\% & 41.17\% & \textbf{53.10\%} \\
\bottomrule
\end{tabular}%
}
\end{table}

\subsubsection{Inference Efficiency and Acceleration}

As shown in \cref{tab:efficiency-comparison}, DualComp achieves the best overall performance while also delivering the highest efficiency. Unlike competing methods, which all use a fixed $24\times$ compression ratio, DualComp reaches $42.4\times$, reducing the average visual token volume from $14.0$k to $6.4$k, and lowering LLM computation from $198.06$ TFLOPs to $99.75$ TFLOPs.

More importantly, DualComp achieves the fastest end-to-end inference without sacrificing accuracy. It runs at $3.87$ s/image, significantly faster than VisionZip ($8.41$ s/image), FastV ($19.56$ s/image), and SparseVLM ($7.84$ s/image), while still achieving the best overall accuracy ($53.10\%$). This speedup is achieved through a lightweight visual compression stage ($1.52$ s) and a shorter LLM generation stage ($2.25$ s).

Unlike methods such as FastV and SparseVLM, which perform token compression after visual tokens are passed to the LLM, DualComp reduces and redistributes visual evidence before the generation phase. VisionZip adopts a hybrid strategy, preserving high-value tokens while merging the remaining ones based on similarity, but it incurs significant visual-side overhead, especially with larger images. In contrast, DualComp not only compresses more aggressively but also achieves the highest performance with the lowest runtime.

\subsection{Ablation Study}

To evaluate the contribution of each component in DualComp, we conduct ablations on the dual-stream design, explicit topology completion, topological sequence unrolling, and text-aware structural modulation.

\begin{table*}[b!]
\footnotesize
\centering
\caption{\textbf{Ablation results of DualComp on XLRS-Bench.}
We compare the full model with variants that remove or modify key components, including \emph{SCSA-only} (w/o geometric stream), \emph{IGSR-only} (w/o semantic stream), \emph{Top-K} (w/o topology), \emph{TASM-off} (w/o text-aware modulation), and \emph{Index-Reorder} (w/o topological unrolling). The subtask abbreviations follow Table~\ref{tab:main-result}.}
\label{tab:ablation}
\resizebox{\textwidth}{!}{%
\begin{tabular}{l|cccccccc|ccccc|c}
\toprule
\rowcolor[gray]{.85}
\textbf{Method} & \multicolumn{8}{c|}{\textbf{Perception}} & \multicolumn{5}{c|}{\textbf{Reasoning}} & \textbf{Avg.} \\
\midrule
\rowcolor[gray]{.95}
\textbf{Sub-tasks} & \textbf{OC} & \textbf{RC} & \textbf{OSR} & \textbf{OLUC} & \textbf{RLUC} & \textbf{OCC} & \textbf{OCL} & \textbf{OMS} & \textbf{AD} & \textbf{ECR} & \textbf{RP} & \textbf{RCCD} & \textbf{CCR} & \\
\midrule
\rowcolor[gray]{.97}
\emph{SCSA-only}                    & 25.0 & 44.0 & 36.0 & 50.0 & 69.0 & 40.0 & 26.5 & \textbf{65.0} & 62.0 & 77.0 & 67.0 & 45.0 & 47.0 & 50.3 \\
\emph{IGSR-only}                    & 25.0 & 42.0 & 34.8 & 50.0 & 68.0 & 43.4 & 28.6 & \textbf{65.0} & 65.0 & 78.0 & 71.0 & 45.0 & 48.0 & 51.1 \\
\rowcolor[gray]{.97}
\midrule
\emph{Top-K}                        & 25.0 & 41.0 & 34.4 & 52.0 & 65.0 & 38.9 & 31.3 & \textbf{65.0} & 65.0 & 76.0 & 65.0 & 45.0 & 50.0 & 50.5 \\
\midrule
\emph{TASM-off}                     & 25.0 & 44.0 & 35.8 & 48.0 & 66.5 & 42.8 & 31.8 & \textbf{65.0} & 67.0 & 77.0 & 71.0 & 45.0 & 47.0 & 51.2 \\
\rowcolor[gray]{.97}
\midrule
\emph{Index-Reorder}                & 25.0 & 42.0 & 35.4 & 51.0 & 66.0 & 43.9 & 32.1 & \textbf{65.0} & 67.0 & 77.0 & \textbf{72.0} & 45.0 & 48.0 & 51.5 \\
\midrule
\rowcolor{rowblue}
\textbf{DualComp}                 & \textbf{26.7} & \textbf{45.0} & \textbf{37.4} & \textbf{53.0} & \textbf{69.5} & \textbf{43.6} & \textbf{34.0} & \textbf{65.0} & \textbf{69.0} & \textbf{79.0} & \textbf{72.0} & 48.3 & 49.0 & \textbf{53.1} \\
\bottomrule
\end{tabular}%
}
\end{table*}

\subsubsection{Dual-Stream Architecture}

We first compare the full model with \emph{SCSA-only (w/o geometric stream)} and \emph{IGSR-only (w/o semantic stream)}. The full model achieves the best overall accuracy ($53.1\%$), outperforming SCSA-only ($50.27\%$) and IGSR-only ($51.06\%$), confirming the complementarity between the semantic and geometric streams. As illustrated in \cref{tab:ablation}, this complementarity is consistently reflected across geometric-dominant, semantically--geometrically balanced, and semantic-dominant tasks.

Removing the geometric stream mainly hurts geometric-dominant tasks. 
For example, RP drops from $72.00\%$ to $67.00\%$, and AD decreases from $69.00\%$ to $62.00\%$, indicating that semantic aggregation alone cannot adequately preserve topology-critical evidence for scene-level reasoning.
In contrast, removing the semantic stream mainly affects semantic-dominant tasks. 
For instance, RC falls from $45.00\%$ to $42.00\%$, and OC decreases from $26.70\%$ to $25.00\%$, showing that geometric recovery alone is insufficient for preserving object-level semantic cues under compression.

The complementarity is even more evident on semantically--geometrically balanced tasks that depend on both fine-grained object semantics and structural context. On OCC, the full model achieves $43.63\%$, outperforming both SCSA-only ($40.00\%$) and IGSR-only ($43.38\%$). 
These results show that neither stream alone is sufficient to cover the full range of semantic and geometric demands in UHR remote sensing, and that the best performance is achieved only when both streams are jointly preserved.

\subsubsection{Explicit Topology Completion}

To assess the role of explicit topology recovery, we construct \emph{Top-K (w/o topology)}, which keeps only the top-scoring structural tokens without path connection. This variant yields the lowest overall accuracy among all ablations ($49.96\%$). As shown in \cref{tab:ablation}, the largest degradation appears on RP, which drops sharply from $72.00\%$ to $60.00\%$; RLUC also decreases from $69.50\%$ to $65.00\%$. This confirms that retaining isolated structural tokens is insufficient for geometric reasoning, and that explicit topology completion is critical for preserving connectivity under aggressive compression.

\subsubsection{Topological Sequence Unrolling}

To evaluate the effect of sequence organization, we construct \emph{Index-Reorder (w/o topological unrolling)}, which preserves the topology paths recovered by greedy path tracing but reorders the selected tokens by their original spatial indices before feeding them into the LLM. \cref{tab:ablation} shows that its overall accuracy drops from $53.1\%$ to $51.49\%$. Although it remains clearly better than \emph{Top-K (w/o topology)} ($51.49\%$ vs. $49.96\%$), it still underperforms the full model, indicating that performance depends not only on recovering the right structural tokens, but also on organizing them in a topology-consistent order. This verifies the contribution of topological sequence unrolling in improving geometric continuity modeling.

\subsubsection{Text-Aware Structural Modulation}

Finally, we construct \emph{TASM-off (w/o text-aware modulation)}, which removes text--vision similarity modulation and relies only on local feature differences to build the structural cost field. As shown in the \cref{tab:ablation}, the overall accuracy decreases from $53.1\%$ to $51.22\%$, showing that text priors further improve the alignment between structure selection and task intent. The drop is more evident on scene understanding tasks, e.g., OLUC decreases from $53.00\%$ to $48.00\%$ and RLUC from $69.50\%$ to $66.50\%$. Still, TASM-off remains stronger than several other ablated variants, suggesting that text-aware modulation is an important enhancement rather than a prerequisite for geometric recovery.


\subsection{Further Analysis: Transferability to General-Purpose MLLMs}

\begin{table*}[t!]
\footnotesize
\centering
\caption{\textbf{Performance comparison on XLRS-Bench based on Qwen2.5-VL-7B.}}
\vspace{-3mm}
\label{tab:qwen-result}
\resizebox{\textwidth}{!}{%
\begin{tabular}{l|c|cccccccc|ccccc|c}
\toprule
\rowcolor[gray]{.85}
\textbf{Method} & \textbf{Compression} & \multicolumn{8}{c|}{\textbf{Perception}} & \multicolumn{5}{c|}{\textbf{Reasoning}} & \textbf{Avg.} \\
\midrule
\rowcolor[gray]{.95}
\textbf{Sub-tasks} & & \textbf{OC} & \textbf{RC} & \textbf{OSR} & \textbf{OLUC} & \textbf{RLUC} & \textbf{OCC} & \textbf{OCL} & \textbf{OMS} & \textbf{AD} & \textbf{ECR} & \textbf{RP} & \textbf{RCCD} & \textbf{CCR} & \\
\midrule
Qwen2.5-VL-7B        & $-$          & 33.3 & 40.0 & 36.2 & 31.0 & 77.0 & 40.6 & 40.5 & 66.7 & 68.0 & 72.0 & 27.0 & 38.3 & 45.0 & 47.4 \\
\midrule
\rowcolor[gray]{.97}
\emph{+ VisionZip}   & $1.43\times$ & 35.0 & 38.0 & 36.8 & 33.0 & 78.5 & 39.0 & 40.6 & 66.7 & 68.0 & 73.0 & 25.0 & 38.3 & 46.0 & 47.5 \\
\emph{+ VisionZip}   & $10\times$   & 36.7 & 42.0 & 34.0 & 33.0 & 78.5 & 38.6 & 40.3 & 66.7 & 60.0 & 72.0 & 24.0 & 38.3 & 46.0 & 46.9 \\
\midrule
\rowcolor{rowblue}
\emph{\textbf{+ Ours}}      & \normalsize \textcolor{darkred}{$\mathbf{10.24\times}$} & \textbf{38.3} & 36.0 & 35.4 & \textbf{36.0} & 72.5 & \textbf{41.9} & \textbf{40.6} & \textbf{66.7} & \textbf{71.0} & \textbf{74.0} & \textbf{29.0} & \textbf{38.3} & 43.0 & \textbf{47.9} \\
\bottomrule
\end{tabular}%
}
\end{table*}

To further evaluate the generality of DualComp, we transplant it to the general-purpose model Qwen2.5-7B. As shown in \cref{tab:qwen-result}, DualComp improves the overall accuracy from $47.40\%$ to $47.90\%$, showing that the proposed framework remains effective beyond remote-sensing-specific backbones. This gain reflects more than backbone compatibility: it suggests that the semantic--geometric duality modeled by DualComp is intrinsic to UHR remote sensing tasks themselves, and therefore remains beneficial even on a general-purpose MLLM.

Under the same $10\times$ compression ratio, DualComp also outperforms VisionZip-$10\times$ ($47.90\%$ vs. $46.92\%$), with especially clear gains on AD ($71\%$ vs. $60\%$), ECR ($74\%$ vs. $72\%$), and RP ($29\%$ vs. $24\%$). Moreover, even compared with a VisionZip variant using a smaller compression ratio, DualComp still achieves higher overall accuracy ($47.90\%$ vs. $47.53\%$), indicating that its advantage comes from more effective semantic--geometric scheduling rather than from a looser compression setting. Overall, these results show that DualComp can be effectively adapted to general-purpose MLLMs while maintaining stable gains on UHR remote sensing tasks.

\section{Conclusion}
In this paper, we target a key bottleneck in scaling MLLMs to ultra-high-resolution (UHR) remote-sensing imagery: the prohibitive inference cost from visual token explosion and the mismatch of static compression policies to task-heterogeneous interpretation. Through a pilot study, we reveal a pronounced semantic–geometric duality: semantic understanding can benefit from background denoising, whereas geometric reasoning depends on preserving background context, structural continuity, and topology as critical evidence. To address this, we propose DualComp, a task-intent-aware dual-stream token compression framework: the semantic stream uses SCSA to compress redundant background while retaining object-related evidence, and the geometric stream employs IGSR to recover structural and connectivity cues under high compression for topology-sensitive reasoning. The compression modules are parameter-free and training-free at deployment, and the router is pretrained offline and frozen at inference, enabling plug-and-play integration without updating host MLLM weights. Extensive results on XLRS-Bench show that DualComp substantially reduces tokens and end-to-end latency while improving accuracy across both semantic- and geometry-dominant tasks, validating the effectiveness of task-aware compression for UHR remote-sensing understanding.
\bibliographystyle{splncs04}
\bibliography{main}

@String(CVPR  = {IEEE Conf. Comput. Vis. Pattern Recog.})

@String(ICCV  = {Int. Conf. Comput. Vis.})

@String(ECCV  = {Eur. Conf. Comput. Vis.})

@String(NeurIPS = {Adv. Neural Inform. Process. Syst.})

@String(ICML  = {Int. Conf. Mach. Learn.})

@String(AAAI  = {AAAI})

@String(CVPR  = {CVPR})

@String(ICCV  = {ICCV})

@String(ECCV  = {ECCV})

@String(NeurIPS = {NeurIPS})

@String(ICML  = {ICML})

@article{gemini,
  title={Gemini: A Family of Highly Capable Multimodal Models},
  author={Gemini Team and Rohan Anil and Sebastian Borgeaud and Jean-Baptiste Alayrac and Jiahui Yu and Radu Soricut and Johan Schalkwyk and Andrew M. Dai and Anja Hauth and Katie Millican and others},
  journal={arXiv preprint arXiv:2312.11805},
  year={2023}
}

@article{gpt-4,
  title={GPT-4 Technical Report},
  author={Josh Achiam and Steven Adler and Sandhini Agarwal faces and Lama Ahmad and Ilge Akkaya and Florencia Leoni Aleman and Diogo Almeida and Janko Altenschmidt and Sam Altman and Shyamal Anadkat and others},
  journal={arXiv preprint arXiv:2303.08774},
  year={2023}
}

@article{llama,
  title={Llama: Open and Efficient Foundation Language Models},
  author={Hugo Touvron and Thibaut Lavril and Gautier Izacard and Xavier Martinet and Marie-Anne Lachaux and Timoth{\'e}e Lacroix and Baptiste Rozi{\`e}re and Naman Goyal and Eric Hambro and Faisal Azhar and others},
  journal={arXiv preprint arXiv:2302.13971},
  year={2023}
}

@article{minigpt,
  title={MiniGPT-4: Enhancing Vision-Language Understanding with Advanced Large Language Models},
  author={Deyao Zhu and Jun Chen and Xiaoqian Shen and Xiang Li and Mohamed Elhoseiny},
  journal={arXiv preprint arXiv:2304.10592},
  year={2023}
}

@article{qwenvl,
  title={Qwen-VL: A Frontier Large Vision-Language Model with Versatile Abilities},
  author={Jinze Bai and Shuai Bai and Shusheng Yang and Shijie Wang and Sinan Tan and Peng Wang and Junyang Lin and Chang Zhou and Jingren Zhou},
  journal={arXiv preprint arXiv:2308.12966},
  volume={1},
  number={2},
  pages={3},
  year={2023}
}

@inproceedings{geochat,
  title={GeoChat: Grounded Large Vision-Language Model for Remote Sensing},
  author={Kartik Kuckreja and Muhammad Sohail Danish and Muzammal Naseer and Abhijit Das and Salman Khan and Fahad Shahbaz Khan},
  booktitle={CVPR},
  pages={27831--27840},
  year={2024}
}

@inproceedings{lhrs,
  title={LHRS-Bot: Empowering Remote Sensing with VGI-Enhanced Large Multimodal Language Model},
  author={Dilxat Muhtar and Zhenshi Li and Feng Gu and Xueliang Zhang and Pengfeng Xiao},
  booktitle={ECCV},
  pages={440--457},
  year={2024}
}

@article{rsgpt,
  title={RSGPT: A Remote Sensing Vision Language Model and Benchmark},
  author={Yuan Hu and Jianlong Yuan and Congcong Wen and Xiaonan Lu and Yu Liu and Xiang Li},
  journal={ISPRS Journal of Photogrammetry and Remote Sensing},
  volume={224},
  pages={272--286},
  year={2025}
}

@misc{llavanext,
  title={LLaVA-NeXT: Improved reasoning, OCR, and world knowledge},
  author={Haotian Liu and Chunyuan Li and Yuheng Li and Bo Li and Yuanhan Zhang and Sheng Shen and Yong Jae Lee},
  year={2024}
}

@article{fargpt-4v,
  title={How far are we to GPT-4V? Closing the gap to commercial multimodal models with open-source suites},
  author={Zhe Chen and Weiyun Wang and Hao Tian and Shenglong Ye and Zhangwei Gao and Erfei Cui and Wenwen Tong and Kongzhi Hu and Jiapeng Luo and Zheng Ma and others},
  journal={Science China Information Sciences},
  volume={67},
  number={12},
  pages={220101},
  year={2024}
}

@inproceedings{prumerge,
  title={LLaVA-PruMerge: Adaptive Token Reduction for Efficient Large Multimodal Models},
  author={Yuzhang Shang and Mu Cai and Bingxin Xu and Yong Jae Lee and Yan Yan},
  booktitle={ICCV},
  pages={22857--22867},
  year={2025}
}

@inproceedings{visionzip,
  title={VisionZip: Longer is Better but Not Necessary in Vision Language Models},
  author={Senqiao Yang and Yukang Chen and Zhuotao Tian and Chengyao Wang and Jingyao Li and Bei Yu and Jiaya Jia},
  booktitle={CVPR},
  pages={19792--19802},
  year={2025}
}

@article{vscan,
  title={VScan: Rethinking Visual Token Reduction for Efficient Large Vision-Language Models},
  author={Ce Zhang and Kaixin Ma and Tianqing Fang and Wenhao Yu and Hongming Zhang and Zhisong Zhang and Yaqi Xie and Katia Sycara and Haitao Mi and Dong Yu},
  journal={arXiv preprint arXiv:2505.22654},
  year={2025}
}

@inproceedings{fastv,
  title={An Image is Worth 1/2 Tokens After Layer 2: Plug-and-Play Inference Acceleration for Large Vision-Language Models},
  author={Liang Chen and Haozhe Zhao and Tianyu Liu and Shuai Bai and Junyang Lin and Chang Zhou and Baobao Chang},
  booktitle={ECCV},
  pages={19--35},
  year={2024}
}

@article{pdrop,
  title={PyramidDrop: Accelerating Your Large Vision-Language Models via Pyramid Visual Redundancy Reduction},
  author={Long Xing and Qidong Huang and Xiaoyi Dong and Jiajie Lu and Pan Zhang and Yuhang Zang and Yuhang Cao and Conghui He and Jiaqi Wang and Feng Wu and others},
  journal={arXiv preprint arXiv:2410.17247},
  year={2024}
}

@inproceedings{FitPrune,
  title={Fit and Prune: Fast and Training-Free Visual Token Pruning for Multi-Modal Large Language Models},
  author={Weihao Ye and Qiong Wu and Wenhao Lin and Yiyi Zhou},
  booktitle={AAAI},
  volume={39},
  number={21},
  pages={22128--22136},
  year={2025}
}

@inproceedings{atp-llava,
  title={ATP-LLaVA: Adaptive Token Pruning for Large Vision Language Models},
  author={Xubing Ye and Yukang Gan and Yixiao Ge and Xiao-Ping Zhang and Yansong Tang},
  booktitle={CVPR},
  pages={24972--24982},
  year={2025}
}

@article{instructblip,
  title={InstructBLIP: Towards General-Purpose Vision-Language Models with Instruction Tuning},
  author={Wenliang Dai and Junnan Li and Dongxu Li and Anthony Tiong and Junqi Zhao and Weisheng Wang and Boyang Li and Pascale N. Fung and Steven Hoi},
  journal={NeurIPS},
  volume={36},
  pages={49250--49267},
  year={2023}
}

@article{mplug,
  title={mPLUG-Owl: Modularization Empowers Large Language Models with Multimodality},
  author={Qinghao Ye and Haiyang Xu and Guohai Xu and Jiabo Ye and Ming Yan and Yiyang Zhou and Junyang Wang and Anwen Hu and Pengcheng Shi and Yaya Shi and others},
  journal={arXiv preprint arXiv:2304.14178},
  year={2023}
}

@article{flamingo,
  title={Flamingo: a Visual Language Model for Few-Shot Learning},
  author={Jean-Baptiste Alayrac and Jeff Donahue and Pauline Luc and Antoine Miech and Iain Barr and Yana Hasson and Karel Lenc and Arthur Mensch and Katherine Millican and Malcolm Reynolds and others},
  journal={NeurIPS},
  volume={35},
  pages={23716--23736},
  year={2022}
}

@article{geollava8k,
  title={GeoLLaVA-8K: Scaling Remote-Sensing Multimodal Large Language Models to 8k Resolution},
  author={Fengxiang Wang and Mingshuo Chen and Yueying Li and Di Wang and Haotian Wang and Zonghao Guo and Zefan Wang and Boqi Shan and Long Lan and Yulin Wang and others},
  journal={arXiv preprint arXiv:2505.21375},
  year={2025}
}

@inproceedings{lrsvqa,
  title={When Large Vision-Language Model Meets Large Remote Sensing Imagery: Coarse-to-Fine Text-Guided Token Pruning},
  author={Junwei Luo and Yingying Zhang and Xue Yang and Kang Wu and Qi Zhu and Lei Liang and Jingdong Chen and Yansheng Li},
  booktitle={ICCV},
  pages={9206--9217},
  year={2025}
}

@article{zoomearth,
  title={ZoomEarth: Active Perception for Ultra-High-Resolution Geospatial Vision-Language Tasks},
  author={Ruixun Liu and Bowen Fu and Jiayi Song and Kaiyu Li and Wanchen Li and Lanxuan Xue and Hui Qiao and Weizhan Zhang and Deyu Meng and Xiangyong Cao},
  journal={arXiv preprint arXiv:2511.12267},
  year={2025}
}

@inproceedings{xlrsbench,
  title={XLRS-Bench: Could Your Multimodal LLMs Understand Extremely Large Ultra-High-Resolution Remote Sensing Imagery?},
  author={Fengxiang Wang and Hongzhen Wang and Zonghao Guo and Di Wang and Yulin Wang and Mingshuo Chen and Qiang Ma and Long Lan and Wenjing Yang and Jing Zhang and others},
  booktitle={CVPR},
  pages={14325--14336},
  year={2025}
}

@article{attention,
  title={Attention Is All You Need},
  author={Ashish Vaswani and Noam Shazeer and Niki Parmar and Jakob Uszkoreit and Llion Jones and Aidan N. Gomez and {\L}ukasz Kaiser and Illia Polosukhin},
  journal={NeurIPS},
  volume={30},
  year={2017}
}

@article{llava-onevision,
  title={LLaVA-OneVision: Easy Visual Task Transfer},
  author={Bo Li and Yuanhan Zhang and Dong Guo and Renrui Zhang and Feng Li and Hao Zhang and Kaichen Zhang and Peiyuan Zhang and Yanwei Li and Ziwei Liu and others},
  journal={arXiv preprint arXiv:2408.03326},
  year={2024}
}

@article{qwen3vl,
  title={Qwen3-VL Technical Report},
  author={Shuai Bai and Yuxuan Cai and Ruizhe Chen and Keqin Chen and Xionghui Chen and Zesen Cheng and Lianghao Deng and Wei Ding and Chang Gao and Chunjiang Ge and others},
  journal={arXiv preprint arXiv:2511.21631},
  year={2025}
}

@article{qwen25vl,
  title={Qwen2.5-VL Technical Report},
  author={Shuai Bai and Keqin Chen and Xuejing Liu and Jialin Wang and Wenbin Ge and Sibo Song and Kai Dang and Peng Wang and Shijie Wang and Jun Tang and others},
  journal={arXiv preprint arXiv:2502.13923},
  year={2025}
}

@article{internvl2,
  title={Enhancing the Reasoning Ability of Multimodal Large Language Models via Mixed Preference Optimization},
  author={Weiyun Wang and Zhe Chen and Wenhai Wang and Yue Cao and Yangzhou Liu and Zhangwei Gao and Jinguo Zhu and Xizhou Zhu and Lewei Lu and Yu Qiao and others},
  journal={arXiv preprint arXiv:2411.10442},
  year={2024}
}

@article{internvl3,
  title={InternVL3: Exploring Advanced Training and Test-Time Recipes for Open-Source Multimodal Models},
  author={Jinguo Zhu and Weiyun Wang and Zhe Chen and Zhaoyang Liu and Shenglong Ye and Lixin Gu and Yuchen Duan bit and Hao Tian and Weijie Su and Jie Shao and others},
  journal={arXiv preprint arXiv:2504.10479},
  year={2025}
}

@techreport{gpt52,
  title={Introducing GPT-5.2},
  author={{OpenAI}},
  institution={OpenAI},
  year={2025},
  url={https://openai.com/index/introducing-gpt-5-2/}
}

@inproceedings{blip2,
  title={BLIP-2: Bootstrapping Language-Image Pre-training with Frozen Image Encoders and Large Language Models},
  author={Junnan Li and Dongxu Li and Silvio Savarese and Steven Hoi},
  booktitle={ICML},
  pages={19730--19742},
  year={2023}
}

@article{honeybee,
  title={The honeybee as a model for understanding the basis of cognition},
  author={Randolf Menzel},
  journal={Nature Reviews Neuroscience},
  volume={13},
  number={11},
  pages={758--768},
  year={2012}
}

@article{cls,
  title={{[CLS]} Attention is All You Need for Training-Free Visual Token Pruning: Make VLM Inference Faster},
  author={Qizhe Zhang and Aosong Cheng and Ming Lu and Zhiyong Zhuo and Minqi Wang and Jiajun Cao and Shaobo Guo and Qi She and Shanghang Zhang},
  journal={arXiv preprint arXiv:2412.01818v1},
  year={2024}
}

@article{sparsevlm,
  title={SparseVLM: Visual Token Sparsification for Efficient Vision-Language Model Inference},
  author={Yuan Zhang and Chun-Kai Fan and Junpeng Ma and Wenzhao Zheng and Tao Huang and Kuan Cheng and Denis Gudovskiy and Tomoyuki Okuno and Yohei Nakata and Kurt Keutzer and others},
  journal={arXiv preprint arXiv:2410.04417},
  year={2024}
}

@article{tome,
  title={Token Merging: Your ViT But Faster},
  author={Daniel Bolya and Cheng-Yang Fu and Xiaoliang Dai and Peizhao Zhang and Christoph Feichtenhofer and Judy Hoffman},
  journal={arXiv preprint arXiv:2210.09461},
  year={2022}
}

@article{adafv,
  title={AdaFV: Rethinking of Visual-Language alignment for VLM acceleration},
  author={Jiayi Han and Liang Du and Yiwen Wu and Xiangguo Zhou and Hongwei Du and Weibo Zheng},
  journal={arXiv preprint arXiv:2501.09532},
  year={2025}
}

@inproceedings{atp,
  title={ATP-LLaVA: Adaptive Token Pruning for Large Vision Language Models},
  author={Xubing Ye and Yukang Gan and Yixiao Ge and Xiao-Ping Zhang and Yansong Tang},
  booktitle={CVPR},
  pages={24972--24982},
  year={2025}
}

@article{visiontrim,
  title={VisionTrim: Unified Vision Token Compression for Training-Free MLLM Acceleration},
  author={Hanxun Yu and Wentong Li and Xuan Qu and Song Wang and Junbo Chen and Jianke Zhu},
  journal={arXiv preprint arXiv:2601.22674},
  year={2026}
}

@article{llavauhd3,
  title={LLaVA-UHD v3: Progressive Visual Compression for Efficient Native-Resolution Encoding in MLLMs},
  author={Shichu Sun and Yichen Zhang and Haolin Song and Zonghao Guo and Chi Chen and Yidan Zhang and Yuan Yao and Zhiyuan Liu and Maosong Sun},
  journal={arXiv preprint arXiv:2511.21150},
  year={2025}
}

@article{glimpse,
  title={A Glimpse to Compress: Dynamic Visual Token Pruning for Large Vision-Language Models},
  author={Quan-Sheng Zeng and Yunheng Li and Qilong Wang and Peng-Tao Jiang and Zuxuan Wu and Ming-Ming Cheng and Qibin Hou},
  journal={arXiv preprint arXiv:2508.01548},
  year={2025}
}

@article{llava,
  title={Visual Instruction Tuning},
  author={Haotian Liu and Chunyuan Li and Qingyang Wu and Yong Jae Lee},
  journal={NeurIPS},
  volume={36},
  pages={34892--34916},
  year={2023}
}

@article{interns1,
  title={Intern-S1: A Scientific Multimodal Foundation Model},
  author={Lei Bai and Zhongrui Cai and Yuhang Cao and Maosong Cao and Weihan Cao and Chiyu Chen and Haojiong Chen and Kai Chen and Pengcheng Chen and Ying Chen and others},
  journal={arXiv preprint arXiv:2508.15763},
  year={2025}
}

@article{skyeyegpt,
  title={SkyEyeGPT: Unifying Remote Sensing Vision-Language Tasks via Instruction Tuning with Large Language Model},
  author={Yang Zhan and Zhitong Xiong and Yuan Yuan},
  journal={ISPRS Journal of Photogrammetry and Remote Sensing},
  volume={221},
  pages={64--77},
  year={2025}
}

@article{earthgpt,
  title={EarthGPT: A Universal Multimodal Large Language Model for Multisensor Image Comprehension in Remote Sensing Domain},
  author={Wei Zhang and Miaoxin Cai and Tong Zhang and Yin Zhuang and Xuerui Mao},
  journal={IEEE Transactions on Geoscience and Remote Sensing},
  volume={62},
  pages={1--20},
  year={2024}
}

@article{VHM,
  title={VHM: Versatile and Honest Vision Language Model for Remote Sensing Image Analysis},
  author={Chao Pang and Xingxing Weng and Jiang Wu and Jiayu Li and Yi Liu and Jiaxing Sun and Weijia Li and Shuai Wang and Litong Feng and Gui-Song Xia and Conghui He},
  journal={arXiv preprint arXiv:2403.20213},
  year={2024},
  url={https://arxiv.org/abs/2403.20213}
}

@article{RScapret,
  title={Large Language Models for Captioning and Retrieving Remote Sensing Images},
  author={Jo{\~a}o Daniel Silva and Jo{\~a}o Magalh{\~a}es and Devis Tuia and Bruno Martins},
  journal={arXiv preprint arXiv:2402.06475},
  year={2024},
  url={https://arxiv.org/abs/2402.06475}
}

@article{Earthmarker,
  title={EarthMarker: A Visual Prompting Multi-modal Large Language Model for Remote Sensing},
  author={Wei Zhang and Miaoxin Cai and Tong Zhang and Jun Li and Yin Zhuang and Xuerui Mao},
  journal={arXiv preprint arXiv:2407.13596},
  year={2024},
  url={https://arxiv.org/abs/2407.13596}
}

@article{LHRSBotNova,
  title={LHRS-Bot-Nova: Improved Multimodal Large Language Model for Remote Sensing Vision-Language Interpretation},
  author={Zhenshi Li and Dilxat Muhtar and Feng Gu and Xueliang Zhang and Pengfeng Xiao and Guangjun He and Xiaoxiang Zhu},
  journal={arXiv preprint arXiv:2411.09301},
  year={2024},
  url={https://arxiv.org/abs/2411.09301}
}

@article{RSUniVLM,
  title={RSUniVLM: A Unified Vision Language Model for Remote Sensing via Granularity-oriented Mixture of Experts},
  author={Xu Liu and Zhouhui Lian},
  journal={arXiv preprint arXiv:2412.05679},
  year={2024},
  url={https://arxiv.org/abs/2412.05679}
}

@article{EarthMind,
  title={EarthMind: Leveraging Cross-Sensor Data for Advanced Earth Observation Interpretation with a Unified Multimodal LLM},
  author={Yan Shu and Bin Ren and Zhitong Xiong and Danda Pani Paudel and Luc Van Gool and Beg{\"u}m Demir and Nicu Sebe and Paolo Rota},
  journal={arXiv preprint arXiv:2506.01667},
  year={2025},
  url={https://arxiv.org/abs/2506.01667}
}

@article{Ringmogpt,
  title={RingMoGPT: A Unified Remote Sensing Foundation Model for Vision, Language, and Grounded Tasks},
  author={Peijin Wang and Huiyang Hu and Boyuan Tong and Ziqi Zhang and Fanglong Yao and Yingchao Feng and Zining Zhu and Hao Chang and Wenhui Diao and Qixiang Ye and Xian Sun},
  journal={IEEE Transactions on Geoscience and Remote Sensing},
  volume={63},
  pages={1--20},
  year={2025},
  doi={10.1109/TGRS.2024.3510833}
}

@article{EarthVL,
  title={EarthVL: A Progressive Earth Vision-Language Understanding and Generation Framework},
  author={Junjue Wang and Yanfei Zhong and Zihang Chen and Zhuo Zheng and Ailong Ma and Liangpei Zhang},
  journal={arXiv preprint arXiv:2601.02783},
  year={2026},
  url={https://arxiv.org/abs/2601.02783}
}

@article{Earthdial,
  title={EarthDial: Turning Multi-sensory Earth Observations to Interactive Dialogues},
  author={Sagar Soni and Akshay Dudhane and Hiyam Debary and Mustansar Fiaz and Muhammad Akhtar Munir and Muhammad Sohail Danish and Paolo Fraccaro and Watson, Campbell D and Klein, Levente J and Fahad Shahbaz Khan and Salman Khan},
  journal={arXiv preprint arXiv:2412.15190},
  year={2025},
  url={https://arxiv.org/abs/2412.15190}
}

@inproceedings{clip,
  title={Learning Transferable Visual Models from Natural Language Supervision},
  author={Alec Radford and Jong Wook Kim and Chris Hallacy and Aditya Ramesh and Gabriel Goh and Sandhini Agarwal and Girish Sastry and Amanda Askell and Pamela Mishkin and Jack Clark and others},
  booktitle={ICML},
  pages={8748--8763},
  year={2021}
}

@article{imagerag,
  title={ImageRAG: Enhancing Ultra High Resolution Remote Sensing Imagery Analysis with ImageRAG},
  author={Zilun Zhang and Haozhan Shen and Tiancheng Zhao and Zian Guan and Bin Chen and Yuhao Wang and Xu Jia and Yuxiang Cai and Yongheng Shang and Jianwei Yin},
  journal={arXiv preprint arXiv:2411.07688},
  year={2024}
}

@article{vicot,
  title={VICoT-Agent: A Vision-Interleaved Chain-of-Thought Framework for Interpretable Multimodal Reasoning and Scalable Remote Sensing Analysis},
  author={Chujie Wang and Zhiyuan Luo and Ruiqi Liu and Can Ran and Shenghua Fan and Xi Chen and Chu He},
  journal={arXiv preprint arXiv:2511.20085},
  year={2025}
}

@article{qwen2vl,
  title={Qwen2-VL: Enhancing Vision-Language Model's Perception of the World at Any Resolution},
  author={Peng Wang and Shuai Bai and Sinan Tan and Shijie Wang and Zhihao Fan and Jinze Bai and Keqin Chen and Xuejing Liu and Jialin Wang and Wenbin Ge and others},
  journal={arXiv preprint arXiv:2409.12191},
  year={2024}
}

@misc{claude,
  author={Anthropic},
  title={Anthropic AI},
  year={2023},
  url={https://www.anthropic.com}
}

@article{gemini25,
  title={Gemini 2.5: Pushing the Frontier with Advanced Reasoning, Multimodality, Long Context, and Next Generation Agentic Capabilities},
  author={Gheorghe Comanici and Eric Bieber and Mike Schaekermann and Ice Pasupat and Noveen Sachdeva and Inderjit Dhillon and Marcel Blistein and Ori Ram and Dan Zhang and Evan Rosen and others},
  journal={arXiv preprint arXiv:2507.06261},
  year={2025}
}

@inproceedings{dota,
  title={DOTA: A large-scale dataset for object detection in aerial images},
  author={Xia, Gui-Song and Bai, Xiang and Ding, Jian and Zhu, Zhen and Belongie, Serge and Luo, Jiebo and Datcu, Mihai and Pelillo, Marcello and Zhang, Liangpei},
  booktitle={Proceedings of the IEEE conference on computer vision and pattern recognition},
  pages={3974--3983},
  year={2018}
}

@article{semi,
  title={Semi-supervised semantic segmentation in earth observation: The minifrance suite, dataset analysis and multi-task network study},
  author={Castillo-Navarro, Javiera and Le Saux, Bertrand and Boulch, Alexandre and Audebert, Nicolas and Lef{\`e}vre, S{\'e}bastien},
  journal={Machine Learning},
  volume={111},
  number={9},
  pages={3125--3160},
  year={2022},
  publisher={Springer}
}
\end{document}